\documentclass[]{article}

\usepackage[pdftex]{graphicx}
\usepackage{amsmath}
\usepackage{amssymb}
\usepackage{latexsym}
\usepackage{subfig}
\usepackage{verbatim}
\usepackage{hyperref}
\usepackage{wrapfig}

\usepackage{enumitem}
\setlist{nolistsep,leftmargin=*}

\newcommand{\oring}[1]{\breve{#1}}
\newcommand{\nacc}[1]{\tilde{#1}}
\newcommand{\mat}[1]{\mathbf{#1}} 
\newcommand{\Kmminv}{\mat{K}^{-1}_{mm}}

\newcommand{\kiKmmm}[1]{\mat{k}_{#1}^{\top}\Kmminv\mat{m}}
\newcommand{\kiKmmu}[1]{\mat{k}_{#1}^{\top}\Kmminv\mat{u}}
\newcommand{\predVar}[1]{k_{#1#1} - \mat{k}_{#1}^{\top}\Kmminv\mat{k}_{#1}}
\newcommand{\EV}{\mathbb{E}}
\newcommand{\DKL}{\mathbb{D}_\mathrm{KL}}
\newcommand{\dataset}{\mathcal{D}}
\newcommand{\Var}{\mathbb{V}}

\newcommand{\defeq}{\triangleq}
\newcommand{\lgN}{\textrm{ln}\mathcal{N}}
\newcommand{\Normal}{\mathcal{N}}
\usepackage{amsthm}

\newtheorem{theorem}{Theorem}

\begin{document}

\title{Probabilistic Network Metrics:\\Variational Bayesian  Network Centrality}

\author{Harold Soh\\Singapore-MIT Alliance for Research and Technology (SMART)\\
\url{haroldsoh@smart.mit.edu}}
\maketitle
\begin{abstract}
Network metrics form a fundamental part of the network analysis toolbox. Used to quantitatively measure different aspects of the network, these metrics can give insights into the underlying network structure and function. In this work, we connect network metrics to modern probabilistic machine learning. We focus on the centrality metric, which is used a wide variety of applications from web search to gene-analysis. First, we formulate an eigenvector-based Bayesian centrality model for determining node importance. Compared to existing methods, our probabilistic model allows for the assimilation of multiple edge weight observations, the inclusion of priors and the extraction of uncertainties. To enable tractable inference, we develop a variational lower bound (VBC) that is demonstrated to be effective on a variety of networks (two synthetic and five real-world graphs). We then bridge this model to sparse Gaussian processes. The  sparse variational Bayesian centrality Gaussian process (VBC-GP) learns a mapping between node attributes to latent centrality and hence, is capable of predicting centralities from node features and can potentially represent a large number of nodes using only a limited number of inducing inputs. Experiments show that the VBC-GP learns high-quality mappings and compares favorably to a two-step baseline, i.e., a full GP trained on the node attributes and pre-computed centralities. Finally, we present two case-studies using the VBC-GP: first, to ascertain relevant features in a taxi transport network and second, to distribute a limited number of vaccines to mitigate the severity of a viral outbreak.
\end{abstract}

\section{Introduction}
Many real-world systems---from brain structures to the world-wide-web---are elegantly represented and analyzed as graphs or networks, i.e., nodes connected by links or edges. This representation lends itself easily to study using a diverse set of tools developed over the years. A not-insignificant portion of the network analysis toolbox is comprised of \emph{network metrics}~\cite{newman2010networks}---powerful quantitative tools for capturing different aspects of networks---which are particularly useful for exploratory data analysis. 

Among the variety of network metrics proposed, node importance or \textit{centrality} has potentially found the most widespread significance. To give a few examples, centrality measures were used in the identification of gene-disease associations~\cite{Ozgur2008},  important actors in a social network~\cite{borgatti2006identifying}, city travel-hubs~\cite{soh2010weighted}, supernodes in climate networks~\cite{donges2009backbone}, and relevant pages in web search~\cite{brin1998anatomy}. 

In this work, we will take a first-step towards bridging network analysis metrics and machine learning (ML) by introducing a probabilistic ML-version of network centrality. Machine learning methods have already been proven useful in the related areas of community detection~\cite{newman2004finding,psorakis2011overlapping,yang2013community}, link prediction~\cite{libennowell2007,al2006link} and network generation~\cite{PalKnoGha12,kim2012multiplicative}. However, the area of network metrics has thus far remained unexplored by the ML community. Leveraging on ML methods not only allows us to build predictive network metric models, but also naturally incorporate network meta-data, such as node and edge attributes. 

The primary contributions in this work are twofold. In Section \ref{sec:BayesCentralityModel}, we develop a novel Bayesian centrality model. Our model extends the popular eigenvector-based measure by considering centralities as latent attributes that are constrained by the eigenvector centrality relation. Currently, a majority of analyses are performed under the assumption of noise-free graphs. But in reality, ``clean'' graphs are the exception, not the norm. Links are typically estimated from noisy data (e.g., measurements in wet-lab experiments), which lead to inaccurate estimates of network measures including centrality~\cite{Platig2013,Frantz2009,borgatti2006robustness}. In particular, the eigenvector centrality measure undergoes a decay in accuracy (relative to the centrality of the true network) as link errors increase~\cite{borgatti2006robustness}. This issue is important in both theoretical and applied settings, e.g., in neuroscience, centrality scores give us insights into brain organization~\cite{bullmore2012economy} and erroneous scores may falsely identify hub modules.
   
By treating centrality estimation as probabilistic latent variable inference, our Bayesian centrality model directly addresses the problem of uncertainty when inferring node importance and permits principled assimilation of repeated weight observations. It also allows for the incorporation of prior knowledge regarding the centralities and the extraction of uncertainties via the posterior. 
To enable tractable inference, we derive an approximate variational lower-bound that can be optimized either offline or iteratively. 

Another key advantage of the probabilistic formulation is that it allows us to connect centrality analysis to modern probabilistic machine learning methods. This is demonstrated by our second contribution: we extend the variational model in Section \ref{sec:CentralityGPModel} to include information from node attributes via a sparse Gaussian Process (GP)~\cite{Quinonero2005,Titsias2009}. In effect, our model learns a mapping between observed node attributes to the latent centralities. This allows us to represent a potentially large number of nodes using a small number of inducing variables---latent function values at specific input locations that introduce conditional independence between the other latent variables---and to directly predict node centralities from node attributes. 

Experiments on seven different datasets (two synthetic and five real-world networks) demonstrate that centralities inferred by our models are in good agreement with scores computed using the true underlying networks (in Section \ref{sec:ExperimentalResults}). In addition, the learnt mapping produces centralities competitive to that of a two-step method, i.e., a full GP trained on the node-attributes and computed centralities. We demonstrate how VBC-GP can be used in two different case studies. First, we use an automatic relevance determination (ARD) kernel to ascertain feature relevance for a Taxi transportation network. Second, we show how the VBC-GP can be used for limited vaccine distribution to constrain the spread of a virus. Finally, we discuss the advantages of the VBC and VBC-GP, along with  current limitations and future work in Section \ref{sec:Discussion}, and conclude this paper in Section \ref{sec:Conclusion}.

\section{Background: Graphs and Centrality Measures}
\label{sec:background}

In this paper, we work with {networks} or {graphs} that contain {vertices} or {nodes}, connected via {links} or {edges}. More formally, a graph $G \defeq (V,E)$ is a tuple consisting of a set of vertices $V = \{v_1, v_2, \dots, v_{|V|} \}$ and a set of edges $e_{k} = (v_j, v_i, w_{ij}) \in E$. The variable $w_{ij}$ grants to each edge a {weight}, which can represent, for example, probabilities of click-throughs in web-surfing or origin-destination counts in transportation modeling. 

By labeling each vertex with an integer label, we can associate with graph $G$ a {weighted adjacency matrix} $\mathbf{W} = [w_{ij}]$; the ordering of the subscripts in $w_{ij}$ indicates an edge from $j$ to $i$. The networks we deal with can be undirected or directed, with two key assumptions: (i) edge weights are {non-negative} $w_{ij} \geq 0$ and (ii) $G$ is {strongly-connected}, i.e., there exist a directed path between any two vertices $v_i, v_j \in V$. The class of graphs that fall under these two assumptions remains broad, e.g., a social network of friends, an epidemiological contact network or an interlinked group of web-pages. 

\subsection{Centrality in Networks: A Brief Introduction}
As mentioned in the introduction, one of the most significant concepts in network analysis is that of node importance or {centrality}. Since ``importance'' depends on context, a number of centrality measures (e.g., betweenness, closeness) have been developed and we refer readers to \cite{newman2010networks} for more details (and descriptions of other network metrics). Let us associate with a graph $G$ a centrality vector $\mathbf{c} = [c_i]_{i=1}^{|V|}$, representing the importance of each node. 

\paragraph{Degree Centrality} The simplest centrality measure is the {degree}, i.e., the number of edges incident upon a node. When dealing with directed weighted graphs, vertices can have both an in-degree $c^{\textrm{in}}_i = \sum_{j\in N(i)} w_{ij}$ and an out-degree $c^{\textrm{out}}_i = \sum_{j\in N(i)} w_{ji}$, 
where $N(i)$ is a neighborhood function that returns the neighbors of node $i$. An practical examples, the ``citation-count'' metric for estimating scientific paper significance is an in-degree measure and the ``influence'' of actors in organizational structures is an out-degree measure. 

\paragraph{Eigenvector Centrality} While conceptually simple, the degree centrality fails to take into account the ``value'' of linked nodes. It is reasonable that having connections to important nodes increases one's own importance. Let us define the eigenvector centrality of a node to be the weighted sum of the centralities of its neighbors:
\begin{align}
    c^{\mathrm{eig}}_i =  \sum_{j\in N(i)} w_{ij} c^{\mathrm{eig}}_j
    \label{eq:sumOfCentralities}
\end{align}
We see that (\ref{eq:sumOfCentralities}) is a generalization of in-degree where in the latter, links are simply assigned a value of one. The centralities are then found by solving the equation:
\begin{align}
    \lambda_1 \mat{c}^{\mathrm{eig}} = \mat{W}\mat{c}^{\mathrm{eig}}
    \label{eq:eigenvectorEq}
\end{align}
where $\lambda_1$ is the principal eigenvalue and $\mat{c}^{\mathrm{eig}}$ is the associated right eigenvector. There are two notable extensions to the canonical eigenvector centrality: Katz Centrality~\cite{katz1953new} and Google's PageRank~\cite{brin1998anatomy}. We hold off  discussion of PageRank until Section \ref{sec:Discussion}, where we consider future work. 

\paragraph{Katz Centrality} One possible issue concerning eqn. (\ref{eq:sumOfCentralities}) is that nodes can possess zero centrality, i.e., those having no incoming edges or nodes only pointed to by the former. The Katz centrality measure advocates a minimal centrality to each node:
\begin{align}
    c^{\mathrm{katz}}_i =  a \sum_{j\in N(i)} w_{ij} c^{\mathrm{katz}}_j + b
\end{align}
where $a$ and $b$ are positive constants. Typically, $b = 1$, which yields,
\begin{align}\mat{c}^{\mathrm{katz}} = (\mat{I} - a \mat{W})^{-1} \mathbf{1},\end{align} 
where $a$ has become the main parameter of interest. To prevent divergence, $a$ is set below the inverse of the maximal eigenvalue, but is otherwise experimentally tuned. We can adopt the perspective that $a$ and $b$ act as simple ``priors'' on node centralities. Taking this notion further, we can extend eigenvector centrality in a probabilistic manner to account for link errors, as we discuss below.

\section{Bayesian Node Centrality}
\label{sec:BayesCentralityModel}
The general framework here is Bayesian in that we assume the centralities are latent random variables that we want to estimate from noisy observations (given some priors). Consider a dataset $\mathcal{D} = \{d_k\}$ where each sample is a tuple consisting of a node and its observed incoming edges, 
$d = \left( i, \{ \, ( \, j, \hat{w}_{ij}\,) \, \}_{j \in N(i)} \right)$ 
where $i$ is the sampled node and $\{(j, \hat{w}_{ij})\}$ is the set of $i$'s observed neighbors and weights. In contrast with existing centrality methods which typically assume a single weight matrix, we allow for repeated (noisy) observations for a single edge. Using this dataset, we wish to obtain a posterior distribution over the node centralities:
\begin{align}
    p(\mathbf{c}, \lambda | \mathcal{D}) \propto p(\mathcal{D} | \mathbf{c}, \lambda) p(\mathbf{c})p(\lambda)
    \label{eq:BayesCentrality} 
\end{align}

Starting with the centrality and eigenvalue priors, recall from Section \ref{sec:background} that, by assumption, our {underlying} network is strongly-connected with non-negative weights. As such, the weighted adjacency matrix induced by our graph is {irreducible}, allowing us to exercise the Perron-Frobenius theorem~\cite{Perron1907,Frobenius1912}:

\begin{theorem}[\bfseries Perron-Frobenius]
Suppose the matrix $\mat{W}$ is irreducible (the graph is strongly connected), then: 
\begin{enumerate}
\item $\mat{W}$ has a positive real eigenvalue $\lambda_1$ larger (or equal to) in magnitude to all other eigenvalues of $\mat{W}$, $\lambda_1 \ge |\lambda|$.
\item $\lambda_1$ has algebraic and geometric multiplicity 1.
\item Associated with $\lambda_1$ is a positive eigenvector $\mat{p} > 0$ (the Perron vector). Any non-negative eigenvector is a multiple of $\mat{p}$. 
\end{enumerate}
\end{theorem}

The Perron–-Frobenius theorem guarantees that the principal eigenvector $\mat{p}$ has all positive elements and 
$\lambda_1$ is a simple eigenvalue with a single \emph{unique} eigenvector (excluding positive multiples). In other words, to compute the centralities, we have to ensure $\mat{c}$ is positive. We place log-normal priors, with open support $(0,\infty)$, on $\mat{c}$ and $\lambda$ to enforce this constraint and incorporate prior notions of node centralities. To facilitate later connections to Gaussian processes (GP), we parametrize our model using Gaussian random variables $\mat{z} = [z_i]$ where $c_i = \exp (z_i)$\footnote{The exponentiation of a normal r.v. $z\sim \Normal(\mu, \sigma^2)$ yields a log-normal r.v., $\exp(z_i) \sim \lgN(\mu, \sigma^2)$.}. Likewise, $\lambda = \exp(z_\lambda)$, yielding priors with the form:
\begin{align}
    p(\mathbf{c}) & = \prod_{i}^{|V|} \exp(z_i) \\
    p(\lambda) & = \exp(z_\lambda)
\end{align}
where  $p(z_i) = \Normal(\oring{\mu}_{i}^2, \oring{\sigma}_{i}^2)$  and $p(z_\lambda) = \Normal(\oring{\mu}_{\lambda}^2,\oring{\sigma}_{\lambda}^2)$. 
The choice of likelihood function $p(\mathcal{D} | \mathbf{z}, z_\lambda)$ will differ depending on application and the underlying observation process. Here, we derive a variational bound based on the assumption that the difference between the log latent and computed centralities are  normally distributed, which was found empirically to be reasonable under varying edge noise conditions. Let $i = i(k)$ denote the $k$-th observed node, then, 
\begin{align}
p(\mathcal{D} | \mathbf{z}, z_\lambda) & =   \prod_{d \in \mathcal{D}} p(d_k|\mat{z}, z_\lambda) \nonumber\\
p({d}_{k} | \mat{z}, z_\lambda) & = \frac{1}{\sqrt{2 \pi \nacc{\sigma}_{i}^2}} \exp \left\{-\frac{(\zeta_i - z_\lambda - z_i)^2 }{2\nacc{\sigma}_{i}^2} \right\}
\label{eq:logNormalLikelihood}
\end{align}
where $\zeta_i = \log \big(\sum_{j \in N(i)} \hat{w}_{ij} \exp(z_j)\big)$. To simplify our exposition, we point-optimize the node-dependent noise terms $\nacc{\sigma}_{i}$ but a prior can be adopted if necessary. Other likelihoods can be applied to induce different centrality measures and fit within the preceding framework. 
   
Inference with this model can be performed using MCMC, which can be computationally demanding. Otherwise, if only point estimates are required, we could perform maximum likelihood or maximum posteriori inference. In the next section, we derive a variational lower-bound for maximizing $\log p(\mathcal{D})$---a ``middle ground'' approach that will allow us to approximate the posterior distribution, but at a lower computational cost than MCMC. 


\subsection{Variational Lower Bound}
We adopt a mean-field variational approximation using a fully-factorized posterior:
\begin{align}
    q(\mathbf{z},z_\lambda) & = \prod_i^{|V|} \exp(z_i) \exp(z_\lambda) \\
    q(z_i) & = \Normal(\mu_i, \sigma_i^2) \\
    q(z_\lambda) & = \Normal(\mu_\lambda, \sigma_\lambda^2) 
\end{align}
which leads to the variational lower-bound:
\begin{align}
\mathcal{L}_1(q) = & \iint q(\mathbf{z},z_\lambda) \log  \frac{p(\mathcal{D}|\mathbf{z}, \lambda) p(\mathbf{z})p(z_\lambda)}{q(\mathbf{z},z_\lambda)} d\mathbf{z} dz_\lambda \nonumber \\
 = & \EV[\log p(\mathcal{D}|\mathbf{z}, \lambda)] - \DKL[ q(\mathbf{z} \| p(\mathbf{z})] -  \DKL[q(z_\lambda) \| p(z_\lambda)] 
    \label{eq:L1}
\end{align}
The first term is the expectation of the likelihood under $\mat{z}$ and $z_\lambda$,
\begin{align}
&\EV[\log p(\mathcal{D}|\mathbf{z}, \lambda)] = -\frac{1}{2}\sum_{k=1}^{|\dataset|} \log 2\pi\nacc{\sigma}_{i}^2 - 
\sum_{k=1}^{|\dataset|}\frac{\EV[(\zeta_i - z_\lambda - z_i)^2]}{2\nacc{\sigma}_{i}^2}.
\label{eq:EpD}
\end{align}
To resolve the expectation on the RHS, we apply Taylor approximations,
\begin{align}
\EV[(\zeta_i - z_\lambda - z_i)^2] & \approx (\EV[\zeta_i] - \EV[z_\lambda] - \EV[z_i])^2 \label{eq:EJensen} \nonumber\\
& = (\EV[\zeta_i] - \mu_\lambda - \mu_i)^2 \nonumber\\
& \approx (\xi_i - \mu_\lambda - \mu_i)^2 
\end{align}
where $ \xi_i \approx \EV[\zeta_i]$ and we have exploited the linearity of expectations and the assumed independence between the random variables to give,
\begin{align}
\xi_i = & \log \left( \sum_{j \in N(i)} \hat{w}_{ij} \EV[\exp(z_j)] \right) - \frac{\sum_{j \in N(i)} \hat{w}_{ij}^2 \Var[\exp(z_j)]}{2(\sum_{j \in N(i)} \hat{w}_{ij}\EV[\exp(z_j)])^2} 
\label{eq:ExiApprox}
\end{align}
The expectation and variance of $c_i = \exp(z_i)$ are readily computed since $c_i$ is log-normally distributed;
\begin{align}
\EV[\exp(z_i)] & = \exp(\mu_i + \sigma_i^2/2) \label{eq:EVzj} \\
\Var[\exp(z_i)] & = (\exp(\sigma^2_i) - 1)\exp(\mu_i + \sigma_i^2)
\end{align}

The last two terms on the RHS of (\ref{eq:L1}) are the KL-divergence between two normal distributions,
\begin{align}
\mathbb{D}_\text{KL} & (\mathcal{N}_q \| \mathcal{N}_p) = { 1 \over 2 } \bigg( \mathrm{tr} \left( \boldsymbol\Sigma_p^{-1} \boldsymbol\Sigma_q \right) + 
 \left( \boldsymbol\mu_p - \boldsymbol\mu_q\right)^{\rm T} \boldsymbol\Sigma_p^{-1} ( \boldsymbol\mu_p - \boldsymbol\mu_q ) -  d -\ln { |  \boldsymbol \Sigma_q | \over | \boldsymbol\Sigma_p | } \bigg)
\end{align}

\paragraph{Summarized Lower-bound} Finally, combining the terms yields a new approximate lower bound:
\begin{align}
\mathcal{L}_2 = & -\frac{1}{2}\sum_{k=1}^{|\dataset|} \log 2\pi\nacc{\sigma}_{i}^2 - \sum_{k=1}^{|\dataset|}\frac{1}{2\nacc{\sigma}_{i}^2}
\left[ (\xi_i - \mu_\lambda - \mu_i\Big)^2 \right] \nonumber \\
& - \mathbb{D}_\text{KL}(q(\mat{z}) \| p(\mat{z})) 
-\mathbb{D}_\text{KL}(q(z_\lambda) \| p(z_\lambda)).
\end{align}
We can find $q(\mathbf{z}, z_\lambda)$ by maximizing $\mathcal{L}_2$ (or equivalently, minimizing $-\mathcal{L}_2$). In this work, we have used conjugate-gradient optimization, but given large datasets, $\mathcal{L}_2$ can be maximized using mini-batches or stochastic gradient ascent. That said, a normal distribution has to be maintained for each node. In the next section, we describe an extension that maps node attributes to centralities via a latent non-linear function.

\section{A Link to Node Attributes via Sparse Gaussian Processes}
\label{sec:CentralityGPModel}

Recent complex networks models proposed by the machine-learning community have involved node attributes (features) as determinants of edge formation. For example, the Multiplicative Attribute Graph (MAG) model~\cite{kim2012multiplicative} connects nodes probabilistically based on binary attributes and affinity matrices. In \cite{PalKnoGha12}, Palla, Knowles and Ghahramani propose a hierarchical Bayesian model using latent node feature vectors to capture the probability of links.

\begin{figure}
\begin{center}
\includegraphics[width=5cm]{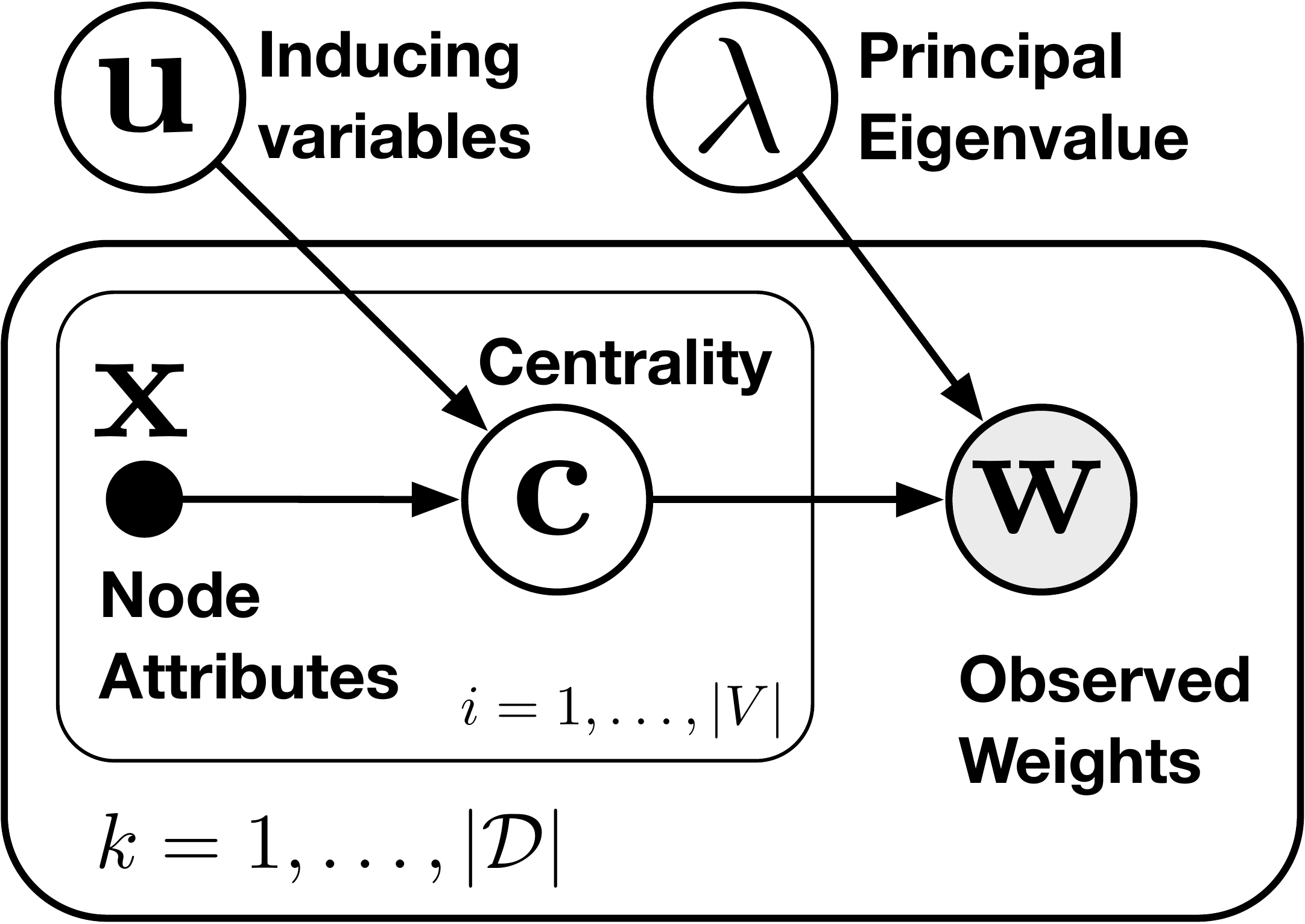}
\end{center}
\caption{Graphical model of the variational Bayesian Centrality Gaussian process (VBC-GP), which uses a sparse GP to learn a mapping from node attributes/features to centralities.}
\label{fig:VBCGP_GraphModel}
\end{figure}

Inspired by this approach, we introduce the variational Bayesian centrality Gaussian Process (VBC-GP), which extends our previous model to consider node attributes as determining factors of centrality in a network. Consider that each node $i$ is associated with an observed feature vector $\mathbf{x}_i$  
and the centralities $\mat{c}$ (and hence, $\mat{z}$) can be well-modelled by a latent function on the node attributes $f(\mathbf{x})$. 
In this work, we use a variational sparse GP~\cite{Titsias2009,Quinonero2005} and introduce the the auxiliary inducing variables $\mat{u}$, the inducing input locations $\mat{\hat{X}} = [\mat{\hat{x}}_i]_{i=1}^m$ and the distributions, 
\begin{align}
p(\mathbf{z} | \mat{u}, \mat{\hat{X}}) & = \mathcal{N}( \mat{K}_{nm} \Kmminv \mat{u} \,,\, \mat{K}_{nn} - \mat{K}_{nm}\Kmminv \mat{K}_{mn}) \nonumber \\
p(\mat{u}) & = \Normal(\mat{0}, \mat{K}_{mm}) \nonumber
\end{align}
where $k_{ij} = k(\mat{x}_i, \mat{x}_j)$, $\mat{k}_i = [k(\mat{x}_i, \mat{\hat{x}}_j)]_{j=1}^{m}$ and $\mat{K}_{mm} = [k(\mat{\hat{x}}_i, \mat{\hat{x}}_j)]_{i,j=1}^{m}$. Adopting the variational distribution, 
\begin{align}
q(\mat{z},\mat{u},z_\lambda) & = p(\mathbf{z} | \mat{u}, \mat{\hat{X}})q(\mat{u})q(z_\lambda) \nonumber \\
q(\mat{u}) & = \mathcal{N}(\mat{m}, \mat{S}), \nonumber 
\end{align}
we formulate a new lower bound, 
\begin{align}
 \mathcal{L}_3(q) &=
 \int p(\mat{z}|\mat{u}, \mat{\hat{X}})q(\mathbf{u},z_\lambda) \log  \frac{p(\mathcal{D}|\mathbf{z}, \lambda) p(\mathbf{u})p(z_\lambda)}{q(\mathbf{u},z_\lambda)} d\mathbf{z} d\mathbf{u} dz_\lambda \nonumber \\
   & = \EV[\log p(\mathcal{D}|\mathbf{z}, \lambda)] - \DKL[ q(\mathbf{u} \| p(\mathbf{u})] - 
     \DKL[q(z_\lambda) \| p(z_\lambda)]. 
    \label{eq:L3}
\end{align}
$\mathcal{L}_3$ bears similarity to $\mathcal{L}_1$, but $\mat{z}$ is now represented indirectly via the sparse GP and the inducing variables $\mat{u}$. This variant enables the prediction of centralities from node features and permits a degree of compression---centralities for a potentially large number of nodes can be represented using a small number of inducing variables.  Figure \ref{fig:VBCGP_GraphModel} shows the complete model (illustrated as a graphical model). 

Since our previous derivations (\ref{eq:L1})-(\ref{eq:EpD}) remain intact, we start from (\ref{eq:EJensen}), 
\begin{align}
\EV[(\zeta_i - z_\lambda - z_i)^2] & \approx (\EV[\zeta_i] - \EV[z_\lambda] - \EV[z_i])^2 \nonumber \\
& = (\EV[\zeta_i] - \mu_\lambda - \kiKmmm{i})^2 \nonumber \\
& \approx (\hat{\xi}_i - \mu_\lambda - \kiKmmm{i})^2 
\end{align} 
where as before, $\hat{\xi}_i \approx \EV[\zeta_i]$ is given by (\ref{eq:ExiApprox}) with expectation and variance, 
\begin{align}
\EV[\exp(z_j)] = & \exp( \kiKmmm{j} + \hat{\sigma}_j^2/2 + \mat{b}^{\top}\mat{S}\mat{b}/2) \label{eq:EVzju}\\
\Var[\exp(z_j)] = & (\exp(\mat{b}^{\top}\mat{S}\mat{b}) -1 )\exp( 2\kiKmmm{j} + \hat{\sigma}_j^2 + \mat{b}^{\top}\mat{S}\mat{b})
\end{align} 
where $\hat{\sigma}_j = \predVar{j}$ (the predictive variance) and $\mat{b} = (\mat{k}_i^{\top}\Kmminv)^{\top}$. In the derivation of the expectation and variance above, we have used the fact that the transformation $\kiKmmu{j} + \hat{\sigma}^2_j/2$ results in a normal distribution $\Normal ( \kiKmmm{j} + \hat{\sigma}_j^2/2 , \mat{b}^\top\mat{S}\mat{b} )$.

\paragraph{Summarized Lower-bound:} Combining the terms results in the approximate lower bound,
\begin{align}
\mathcal{L}_4 = & -\frac{1}{2}\sum_{k=1}^{|\dataset|} \log 2\pi\nacc{\sigma}_{i}^2 - \sum_{k=1}^{|\dataset|}\frac{1}{2\nacc{\sigma}_{i}^2}
\left[ (\hat{\xi}_i - \mu_\lambda - \kiKmmm{i}\Big)^2 \right] \nonumber \\
& - \mathbb{D}_\text{KL}(q(\mat{u}) \| p(\mat{u})) 
-\mathbb{D}_\text{KL}(q(z_\lambda) \| p(z_\lambda)).
\label{eq:VBCGPLowerBound}
\end{align}
We see $\mathcal{L}_4$ is akin to $\mathcal{L}_3$ except that the resulting expectations are additionally taken with respect to $\mat{u}$. Maximizing $\mathcal{L}_4$ yields the approximate posterior $q(\mat{u})q(z_\lambda)$, which can be used to obtain the centrality of a node $\mat{x}_*$, i.e., we exponentiate the normal predictive distribution, which results in a log-normal $c_* \sim \lgN(\kiKmmm{*}, \predVar{*} + \mat{b}^\top\mat{S}\mat{b})$.

\section{Experimental Results}
\label{sec:ExperimentalResults}
In this section, we present results of experiments using VBC and VBC-GP to infer centrality in two synthetic and five real-world networks, i.e., the Erd\H{o}s-R\'{e}nyi (ER) random graph~\cite{erdHos1960evolution} and Bab\'{a}rasi-Albert (BA) scale-free network~\cite{barabasi1999emergence}, a cat brain connectome (CA)~\cite{de2013rich}, the neural network of the \emph{C. elegans} nematode (CE2)~\cite{white1986structure,watts1998collective}, a Facebook user's ego network (FB)~\cite{mcauley2012learning}, Freeman's network of communication between 32 academics (FM)~\cite{freeman1979networkers} and the Singapore public-transportation people-flow network (PT). Basic properties of the networks are given in Table \ref{tbl:GraphProperties}. Source code for the VBC and VBCGP and datasets are available as supplementary material. 

\begin{table}
\caption{Network Dataset Properties. Number of nodes $|V|$, Number of edges $|E|$, attributes, weights, average  degree $\langle k \rangle$ and average clustering coefficient  $\langle C_c \rangle$. }
\label{tbl:GraphProperties}
\begin{center}
\footnotesize
\begin{tabular}{l | c c c c c c}
Network & Nodes $|V|$ & Edges $|E|$ & Attributes & Weights &  Degree $\langle k \rangle$ & Clustering $\langle C_c \rangle$ \\
\hline \\
ER & 100 & 1028 & 0 & N & 10.28 & 0.11 \\ 
BA & 100 & 908 & 0 & N & 9.08 & 0.18 \\ 
CAT & 65 & 1460 & 0 & Y & 22.46 & 0.66 \\ 
CE2 & 239 & 1912 & 0 & N & 8.00 & 0.05 \\ 
FM & 32 & 552 & 3 & Y & 17.25 & 0.86 \\ 
FB & 224 & 6384 & 3 & N & 28.50 & 0.54 \\ 
PT & 57 & 3249 & 2 & Y & 57.00 & 1.00 \\ 
\end{tabular}
\end{center}
\end{table}

\subsection{Centrality with Noisy Edges} 
Our first set of experiments was designed to test if the VBC model was capable of estimating centralities given repeated noisy edge observations. 
For each network, we corrupted the link weights with log-normal noise having variance $\sigma_o^2 = 1.0, 5.0, 10.0$. We provided VBC with an increasing number of samples per node ($N_s = 1, 2, \dots, 10$) and each ``run'' was repeated 15 times. As a performance measure, we use the Kendall correlation to eigenvector centralities of the noise-free network ($\mat{c}_{\mathrm{NF}}$) and the top-10 score (the overlap between the top-10 highest ranking nodes)\footnote{We also computed the Pearson and Spearman correlations. All the scores were strongly-correlated and hence, only the Kendall score is shown here for conciseness.}. The baseline (BL) centralities $\mat{{c}}_{\mathrm{BL}}$, were computed from the averaged weight matrix $\mat{\overline{W}} = \left[N^{-1}_{ij} \sum \hat{w}_{ij}\right]_{i,j=1}^{|V|}$ where $N_{ij}$ is the number of observations for weight $w_{ij}$. The lower-bounds were optimized using a conjugate-gradients method with the initial point set using $\mat{c}_{\mathrm{BL}}$. To simplify the optimization process, we parameter-tied the noise terms $\tilde{\sigma}_i^2 = \tilde{\sigma}_n^2$.

\begin{figure*}
\centering
\includegraphics[width=0.7\textwidth]{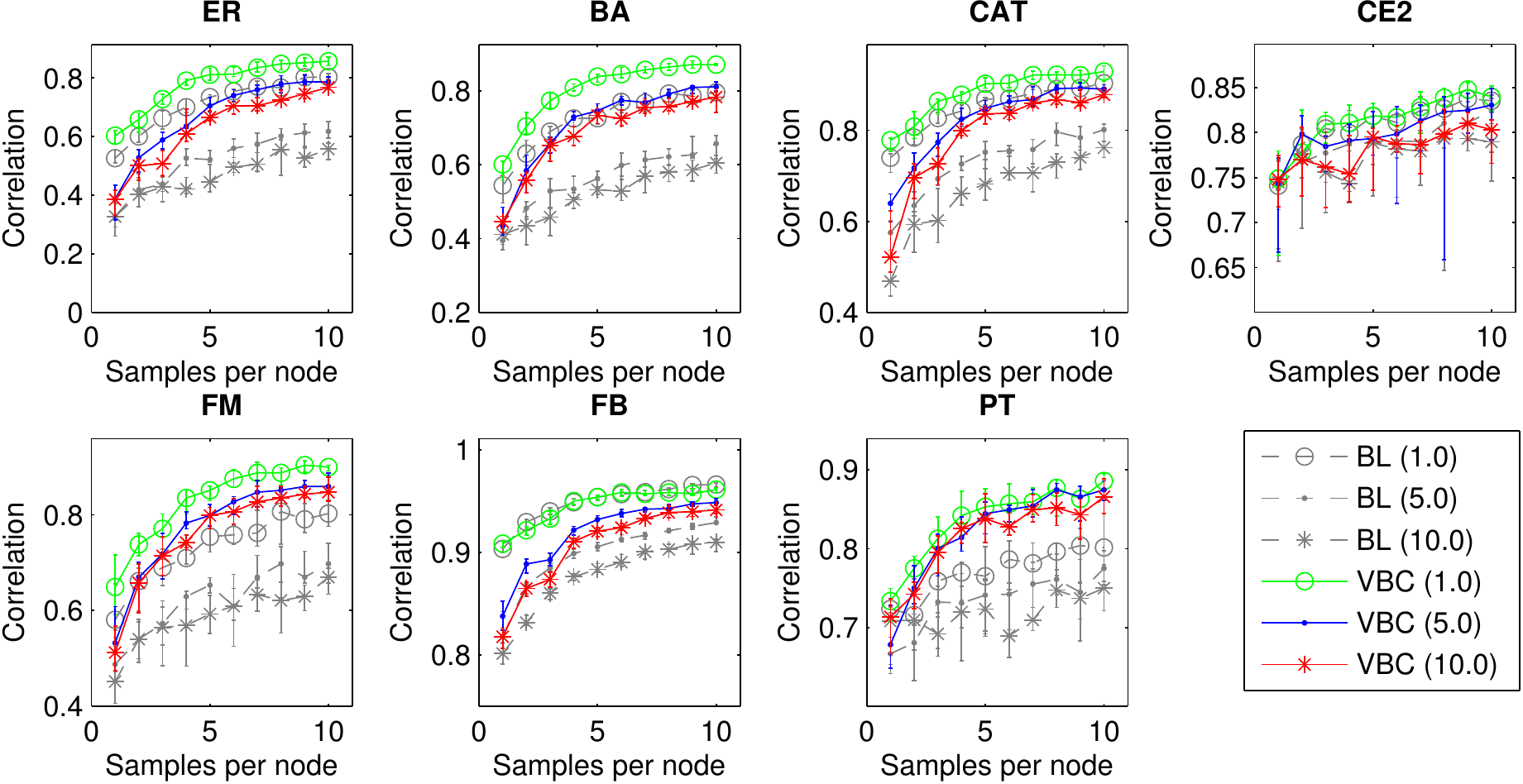}
\caption{Median correlation scores (with upper and lower quartiles) for each of the networks for three noise levels and different sample sizes. The VBC centralities achieved higher correlation scores compared to the baseline (BL) and more quickly approached the noise-free centralities as more samples were provided, particularly for the higher noise levels.}
\label{fig:top10}
\end{figure*}

\begin{figure*}
\centering
\includegraphics[width=0.7\textwidth]{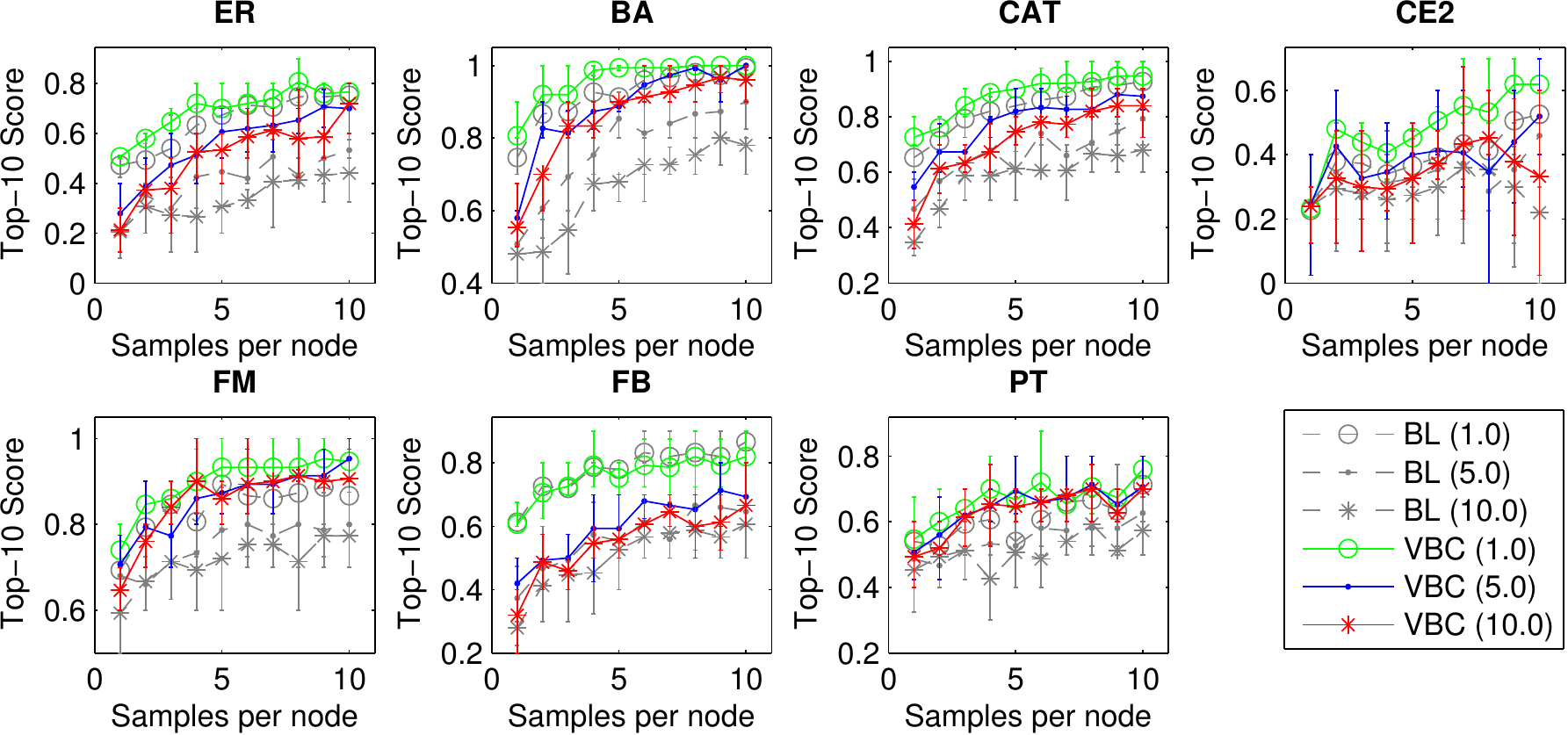}
\caption{Median top-10 overlap scores, with upper and lower quartiles. Similar to the correlation scores, VBC achieved better rankings with more samples.}
\label{fig:convergence}
\end{figure*}

Fig. \ref{fig:convergence} shows that with a single sample per node, the VBC infers centralities comparable to the baseline. With larger $N_s$, the Bayesian centralities become increasingly correlated with the centralities of the true underlying network (rising by $10-20\%$) and at a rate faster than BL. The performance differences are particularly noticeable with noise levels $\sigma_o^2 = 5.0$ and $\sigma_o^2 = 10.0$. The plots for the top-10 scores(Fig. \ref{fig:top10}) are similar, showing that the VBC manages to isolate important nodes better than the baseline. Furthermore, unlike the simple averaging method, the VBC allows for the incorporation of prior knowledge and gives a posterior distribution over the inferred centralities. 

\subsection{Mapping Attributes to Centralities} 
In this experiment, we worked with the FM, FB and PT networks, which possessed node attributes, and generated noisy datasets with link noise $\sigma_o^2 = 5.0$. As before, 15 runs were conducted for each sample size setting $N_s$. The VBC-GP was set to use a squared exponential kernel with an initial lengthscale of $l = 15, 20, 20$ with $m = 10, 20, 40$ inducing inputs for FM, PT and FB respectively. Following \cite{hensman2013}, the inducing input locations $\mathbf{\hat{X}}$ were selected via k-means clustering and the hyperpararameters were optimized along with the other variational parameters. As a baseline model, we trained a full GP with hyperparameter optimization (no sparsity) on the (pre-computed) averaged weight centralities $\mat{c}_\mathrm{BL}$ using 80\% of the nodes. The GP was then used to predict the centralities of all the nodes, which were compared to the true centralities via Kendall correlation. 

\begin{figure}
\centering
\includegraphics[width=0.65\textwidth]{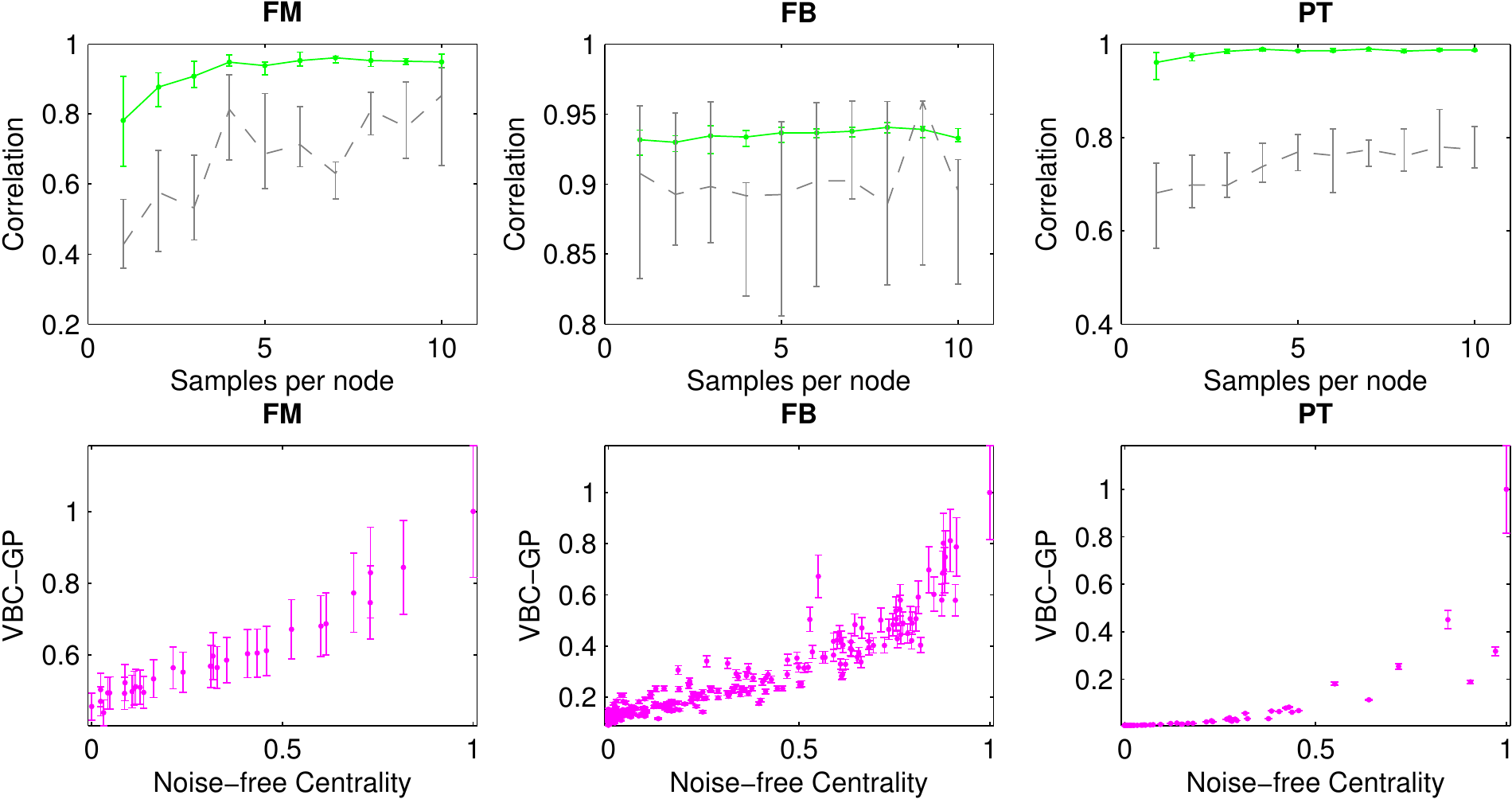}
\caption{The VBC-GP learned high-quality mappings compared to the Full-GP (dashed-line) with higher correlation scores as $N_s$ increased (FM and PT). A decrease in the interquartile ranges suggests more robust performance. Scatter plots (bottom) show the predicted centralities.}
\label{fig:Exp2Results}
\end{figure}

The scores obtained by full-GP (dashed line) and VBC-GP are shown in Fig. \ref{fig:Exp2Results}. Similar to the VBC, the VBC-GP obtained better scores with more samples and outperformed the Full-GP on the FM and PT networks (difference of $7.1$\% and $27.8$\% at $N_s = 10$ respectively). The baseline appeared to slightly outperform the VBC-GP on FB (difference of $0.3$\% at $N_s = 10$). With all three networks, the interquartile ranges for the VBC-GP results were narrower---a tell-tale sign of more robust performance. Overall, the VBC-GP predicted centralities---using only 10 to 40 inducing inputs---were strongly correlated with the true centralities for all three networks, as can be visually confirmed by the sample scatter plots (bottom plots in Fig. \ref{fig:Exp2Results}).

\subsection{Case Study: Relevant Node Features for a Taxi Network}
In this case study, we investigate the relevance of node features for a taxi transportation network. The dataset consists of the number of daily taxi trips between 57 districts in the city-state of Singapore for five weekdays in August 2010. We also obtained one additional dataset of total flows for the entire month and computed ``gold standard'' centralities. To determine which features are relevant to centrality,  we use the automatic relevance determination (ARD) kernel~\cite{neal1996bayesian}:
\begin{align}
	k_{\textrm{ARD}}(\mat{x}, \mat{x}') = \exp{\left\{ (\mat{x} - \mat{x'})\mat{M}(\mat{x} - \mat{x'})  \right\}}
\end{align}
where $\mat{M} = \textrm{diag}([l_{i}^{-2}]_{i = 1}^{d})$ is a diagonal matrix of lengthscales. Each lengthscale $l_k$  determines the impact that an input dimension $k$ has on the output; specifically, the function's responsiveness to input $k$ is inversely related to $l_k$. As such, the smaller hyperparameters indicate which node attributes contribute more to centrality (assuming the ranges are normalized). 

\begin{figure}
\centering
\includegraphics[width=0.7\textwidth]{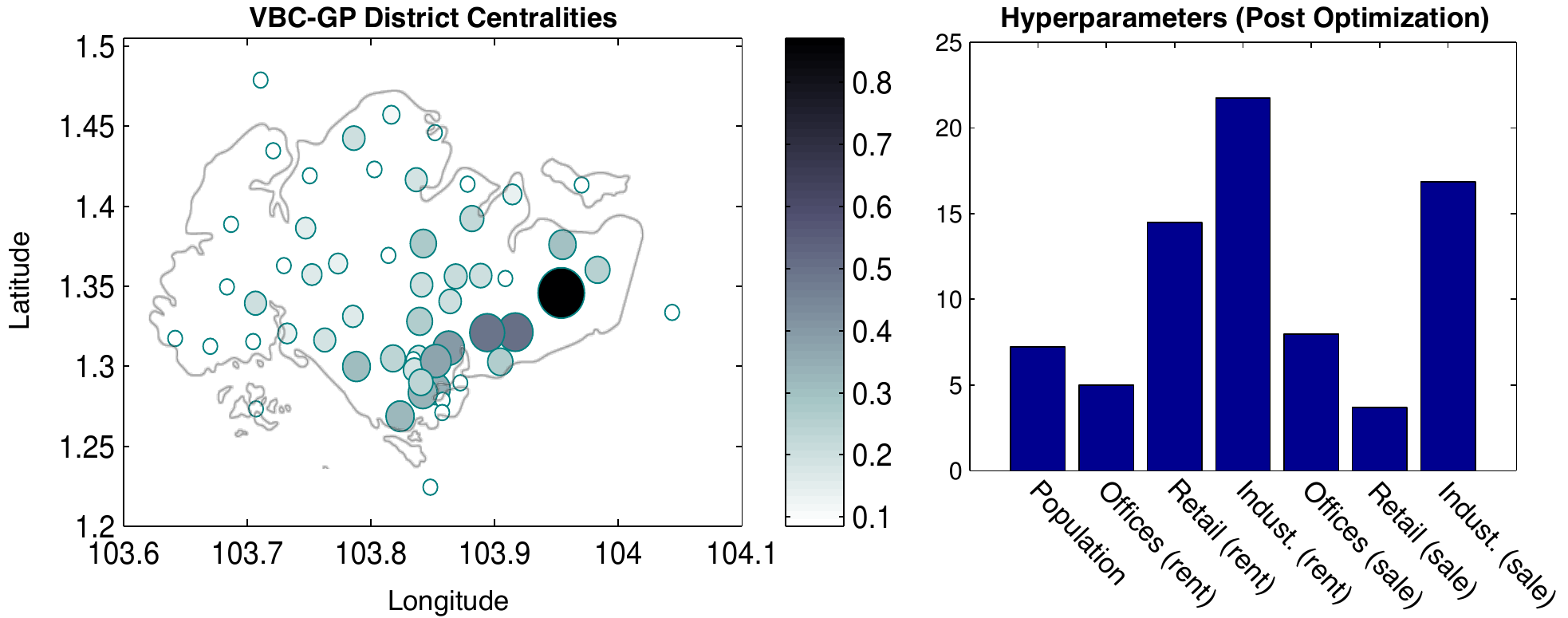}
\caption{(Left) District centroids and associated centralities (size and color) on the taxi network. (Right) Hyperparameters after optimization; the top 3 features were the retail space for sale, office space for rent and resident population.}
\label{fig:ExpTaxiResults}
\end{figure}

We trained the VBC-GP with 40 inducing inputs and equal initial hyperparameters of 50 for a maximum of 1000 iterations. The resultant centralities (geographically plotted) and optimized hyperparameters are shown in Fig. \ref{fig:ExpTaxiResults}. Overall, the centralities obtained are in good agreement with the ``gold standard'' eigenvector centralities---the correlation scores were 0.97 (Pearson) and 0.81 (Kendall) with a top-10 overlap score of 80\% and a top-5 score of 100\%. 

Three most relevant input dimensions were the retail space (for sale), office space (for rent) and total resident population, coinciding with prominent features of the central nodes. The highest ranked district, Tampines, is one of the largest residential areas and commercial districts in Singapore. The second most central node, Bedok, has the largest residential population. Nodes in the commercial district (e.g., the Downtown Core) have low residential populations, but dense office and retail areas, and were also highly ranked. The lowest ranked nodes comprise sparsely-populated and industrial areas. These results highlight that high taxi-traffic nodes comprise major home and office locations and support the intuitive notion that taxi flows on weekdays mainly consist of work-home trips.

\subsection{Case Study: Targeted Vaccine Distribution}
\label{sec:CaseStudy}
How to distribute a limited number of vaccines or prophylactic treatment in a susceptible population remains a challenging and relevant problem in epidemiology. In this case study, we investigate using VBC-GP to identify the top K-central individuals to immunize and potentially slow disease spread. We simulated viral disease outbreaks on contact networks using  discrete-time Susceptible-Infectious-Recovered (SIR) dynamics, i.e., infected agents infect susceptibles who, in turn, may spread the disease before recovering. The simulated disease was set to have a time-independent probability of transmission, $p(I_t) = 0.5$, with recovery rate of 0.1. 

\begin{figure}
\centering
\includegraphics[width=0.70\textwidth]{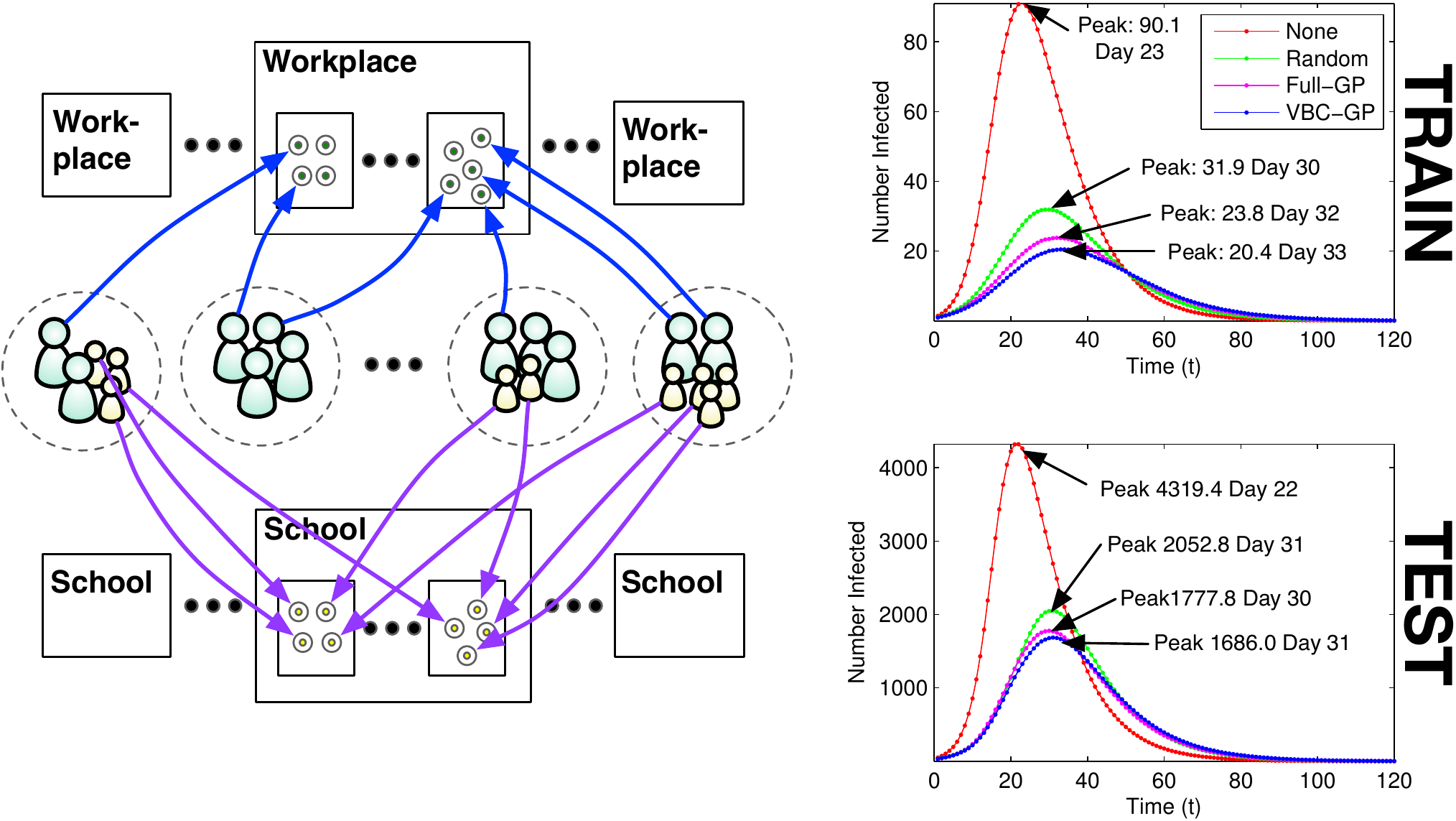}
\caption{(Left) The population contact network generative process where contacts are formed in assigned locations. (Right) Disease progression profiles---averaged number of infected individuals over time for the Train network of $\approx$ 360 people (12,000 edges) and larger Test network of  $\approx$  17,000 people (540,000 edges).}
\label{fig:EpiResults}
\end{figure}
The contact network was generated using location-based hierarchical link formation~\cite{zhang2011temporal}; for example, other than family members, students are assumed to form contacts primarily in classrooms and to a lesser extent, with other students in the same school. Likewise, contacts are probabilistically sampled for workplaces (firms comprising departments), pre-schools, universities and residential communities. The edge weights, $w_{ij} \in [0,1]$ represent the probability of a contact. As such, the virus propagates along a given edge $e_{ij}$ with  probability $w_{ij}p(I_t)$. The observed weights are noisy samples from a truncated normal $\mathcal{N}(w_{ij}, 0.5)$ in the interval $[0,1]$. Each individual was associated with node attributes comprising his/her age and the visited location sizes (which were reduced to 3 dimensions using PCA). Note that the location attributes are \emph{not} the number of contacts formed, e.g., a student may establish only 5 contacts in a class of 20. 

Fig. \ref{fig:EpiResults} (right) illustrates the epidemic profile---averaged over 100 simulation runs on 15 different networks starting with 1\% infected---comparing scenarios with zero vaccination, random baseline, Full-GP baseline and sparse VBC-GP (40 inducing inputs) distribution of vaccines in a population of 360 people (12,000 edges, 30\% vaccination). Treating the VBC-GP selected individuals decreased the total number of infected by 62\% (239.5 to 90.7 people) and the peak (91.0 to 20.4 people and a delay from 23 to 33 days)---better than the baselines. 

To evaluate model generalization, we constructed a larger population of $\approx 17,000$ people and used a VBC-GP model (trained on the smaller networks) to select 30\% of the nodes for vaccination, using only \emph{predicted} node centralities. Fig. \ref{fig:EpiResults} (bottom) shows that the VBC-GP-based vaccine distribution on the novel network (sampled using the same process) resulted in a similar positive outcomes, i.e., a decrease in the total infected by 41\% (from 9983 to 5853 people) and the peak (4319 to 1686 people). 

In summary, immunizing the VBC-GP selected nodes reduced the severity of the outbreaks---a sharp fall in the peak and total infected reduces the strain on healthcare systems. Although our study focusses on a biological virus, similar analyses can be conducted for mitigating the spread of computer viruses or reducing the impact of disruptions/attacks on transportation hubs.

\section{Discussion: Benefits, Limitations and Future Work}
\label{sec:Discussion}
The experimental results demonstrate the merits of our approach, i.e., the ability to handle noisy links and assimilate additional sample observations to better estimate node centralities. The results show that the VBC-GP can learn high-quality mappings in the presence of noise; the correlations compare favorably to the Full-GP, with additional benefits: the model is sparse and the centralities are computed along with the mapping in a single optimization process. 

The proposed model can be improved in several ways. From an optimization standpoint, the derived loss function is non-convex with local-minima and we have found that proper initialization is necessary for good results; better modeling of the noise priors $\tilde{\sigma}_i$ would likely improve performance and future work may look into mechanisms for more robust optimization. Regarding scalability, real-world networks can be very large, comprising  millions of nodes. The VBC-GP---promising given its sparse nature---may be trained using  stochastic variational inference~\cite{hensman2013} for very large networks. 

Another future research topic would be temporal centrality~\cite{kim2012temporal}; in many networks, node importance may change over time. Using an appropriate kernel (e.g., \cite{Soh2014a}), the VBC-GP may capture the dynamics of evolving networks. 

Finally, although the Gaussian likelihood worked well in our experiments, it may be inappropriate where the noise process results in different (perhaps multi-modal) deviation distributions. In such cases, an alternative likelihood function can be substituted into (\ref{eq:BayesCentrality}). Different likelihoods can also be applied to induce different notions of centrality, e.g., PageRank is a variant of eigenvector centrality with centralities weighted by the outgoing links. Incorporating such a weighting into the likelihood model would result in a Bayesian form of PageRank centrality.  Experimentation with different likelihoods under varying noise conditions would make interesting future work.

\section{Conclusion}
\label{sec:Conclusion}
In this paper, we contributed two Bayesian models for eigenvector centrality and derived variational lower-bounds for finding approximate posteriors. The second model incorporates a sparse GP that maps node-attributes to latent centralities, thus alleviating the need to maintain separate distributions for each node. Taking a broader perspective, this paper takes a first-step towards connecting metrics for networks with modern machine learning and we hope that it will catalyze  development of future network metric models. For example, ML-type models may be formulated for alternative centrality definitions (e.g., betweenness) and other metrics such as assortativity and the clustering coefficient. Additionally, while we have used Gaussian processes in this work, future models may feature alternative ML methods.

\bibliographystyle{ieeetr}
\bibliography{varBayesCentrality}

\begin{thebibliography}{10}

\bibitem{newman2010networks}
M.~Newman, {\em {Networks: An Introduction}}.
\newblock Oxford University Press, 2010.

\bibitem{Ozgur2008}
A.~{\"O}zg{\"u}r, T.~Vu, G.~Erkan, and D.~R. Radev, ``Identifying gene-disease
  associations using centrality on a literature mined gene-interaction
  network,'' {\em Bioinformatics}, vol.~24, pp.~i277--i285, 07 2008.

\bibitem{borgatti2006identifying}
S.~P. Borgatti, ``Identifying sets of key players in a social network,'' {\em
  Computational \& Mathematical Organization Theory}, vol.~12, no.~1,
  pp.~21--34, 2006.

\bibitem{soh2010weighted}
H.~Soh, S.~Lim, T.~Zhang, X.~Fu, G.~K.~K. Lee, T.~G.~G. Hung, P.~Di,
  S.~Prakasam, and L.~Wong, ``Weighted complex network analysis of travel
  routes on the singapore public transportation system,'' {\em Physica A},
  vol.~389, no.~24, pp.~5852--5863, 2010.

\bibitem{donges2009backbone}
J.~F. Donges, Y.~Zou, N.~Marwan, and J.~Kurths, ``The backbone of the climate
  network,'' {\em EPL (Europhysics Letters)}, vol.~87, no.~4, p.~48007, 2009.

\bibitem{brin1998anatomy}
S.~Brin and L.~Page, ``The anatomy of a large-scale hypertextual web search
  engine,'' {\em Computer networks and ISDN systems}, vol.~30, no.~1,
  pp.~107--117, 1998.

\bibitem{newman2004finding}
M.~E. Newman and M.~Girvan, ``Finding and evaluating community structure in
  networks,'' {\em Physical Review E}, vol.~69, no.~2, p.~026113, 2004.

\bibitem{psorakis2011overlapping}
I.~Psorakis, S.~Roberts, M.~Ebden, and B.~Sheldon, ``Overlapping community
  detection using bayesian non-negative matrix factorization,'' {\em Physical
  Review E}, vol.~83, no.~6, p.~066114, 2011.

\bibitem{yang2013community}
J.~Yang, J.~McAuley, and J.~Leskovec, ``Community detection in networks with
  node attributes,'' in {\em 2013 IEEE 13th International Conference on Data
  Mining (ICDM)}, pp.~1151--1156, IEEE, 2013.

\bibitem{libennowell2007}
D.~Liben-Nowell and J.~Kleinberg, ``The link-prediction problem for social
  networks,'' {\em Journal of the American Society for Information Science and
  Technology}, vol.~58, no.~7, pp.~1019--1031, 2007.

\bibitem{al2006link}
M.~Al~Hasan, V.~Chaoji, S.~Salem, and M.~Zaki, ``Link prediction using
  supervised learning,'' in {\em SDM’06: Workshop on Link Analysis,
  Counter-terrorism and Security}, 2006.

\bibitem{PalKnoGha12}
K.~Palla, D.~A. Knowles, and Z.~Ghahramani, ``An infinite latent attribute
  model for network data,'' in {\em Proceedings of the 29th International
  Conference on Machine Learning}, July 2012.

\bibitem{kim2012multiplicative}
M.~Kim and J.~Leskovec, ``Multiplicative attribute graph model of real-world
  networks,'' {\em Internet Mathematics}, vol.~8, no.~1-2, pp.~113--160, 2012.

\bibitem{Platig2013}
J.~Platig, E.~Ott, and M.~Girvan, ``Robustness of network measures to link
  errors,'' {\em Phys. Rev. E}, vol.~88, p.~062812, Dec 2013.

\bibitem{Frantz2009}
T.~L. Frantz, M.~Cataldo, and K.~M. Carley, ``Robustness of centrality measures
  under uncertainty: Examining the role of network topology,'' {\em Comput Math
  Organ Theory}, vol.~15, no.~4, pp.~303--328, 2009.

\bibitem{borgatti2006robustness}
S.~P. Borgatti, K.~M. Carley, and D.~Krackhardt, ``On the robustness of
  centrality measures under conditions of imperfect data,'' {\em Social
  networks}, vol.~28, no.~2, pp.~124--136, 2006.

\bibitem{bullmore2012economy}
E.~Bullmore and O.~Sporns, ``The economy of brain network organization,'' {\em
  Nature Reviews Neuroscience}, vol.~13, no.~5, pp.~336--349, 2012.

\bibitem{Quinonero2005}
J.~Qui\~{n}onero Candela and C.~E. Rasmussen, ``A unifying view of sparse
  approximate gaussian process regression,'' {\em Journal of Machine Learning
  Research}, vol.~6, pp.~1939--1959, 2005.

\bibitem{Titsias2009}
M.~Titsias, ``{Variational Learning of Inducing Variables in Sparse Gaussian
  Processes},'' in {\em {The 12th International Conference on Artificial
  Intelligence and Statistics (AISTATS)}}, vol.~5, 2009.

\bibitem{katz1953new}
L.~Katz, ``A new status index derived from sociometric analysis,'' {\em
  Psychometrika}, vol.~18, no.~1, pp.~39--43, 1953.

\bibitem{Perron1907}
O.~Perron, ``Zur theorie der matrices,'' vol.~64, no.~2, pp.~248--263, 1907.

\bibitem{Frobenius1912}
G.~Frobenius, {\em {Uber Matrizen aus nicht negativen Elementen}}.
\newblock Sitzungsberichte K\"{o}niglich Preussichen Akademie der Wissenschaft,
  1912.

\bibitem{erdHos1960evolution}
P.~Erd{\H{o}}s and A.~R{\'e}nyi, ``On the evolution of random graphs,'' {\em
  Publications of the Mathematical Institute of the Hungarian Academy of
  Sciences}, vol.~5, pp.~17--61, 1960.

\bibitem{barabasi1999emergence}
A.-L. Barab{\'a}si and R.~Albert, ``Emergence of scaling in random networks,''
  {\em Science}, vol.~286, no.~5439, pp.~509--512, 1999.

\bibitem{de2013rich}
M.~A. de~Reus and M.~P. van~den Heuvel, ``Rich club organization and
  intermodule communication in the cat connectome,'' {\em J. Neuroscience},
  vol.~33, no.~32, pp.~12929--12939, 2013.

\bibitem{white1986structure}
J.~G. White, E.~Southgate, J.~N. Thomson, and S.~Brenner, ``The structure of
  the nervous system of the nematode caenorhabditis elegans,'' {\em Phil.
  Trans. B}, vol.~314, no.~1165, pp.~1--340, 1986.

\bibitem{watts1998collective}
D.~J. Watts and S.~H. Strogatz, ``Collective dynamics of
  `small-world'networks,'' {\em nature}, vol.~393, no.~6684, pp.~440--442,
  1998.

\bibitem{mcauley2012learning}
J.~J. McAuley and J.~Leskovec, ``Learning to discover social circles in ego
  networks.,'' in {\em NIPS}, vol.~272, pp.~548--556, 2012.

\bibitem{freeman1979networkers}
S.~C. Freeman and L.~C. Freeman, {\em The networkers network: a study of the
  impact of a new communications medium on sociometric structure}.
\newblock Sch. of Soc. Sciences, Uni. of Calif., 1979.

\bibitem{hensman2013}
J.~Hensman, N.~Fusi, and N.~D. Lawrence, ``{Gaussian Processes for Big Data},''
  in {\em Uncertainty in Artificial Intelligence (UAI-13)}, 2013.

\bibitem{neal1996bayesian}
R.~Neal, {\em Bayesian learning for neural networks. \normalfont{Lecture Notes
  in Statistics}}.
\newblock No.~118 in Lecture Notes in Statistics, New York, NY: Springer
  Verlag, 1996.

\bibitem{zhang2011temporal}
T.~Zhang, X.~Fu, C.~K. Kwoh, G.~Xiao, L.~Wong, S.~Ma, H.~Soh, G.~K.~K. Lee,
  T.~Hung, and M.~Lees, ``Temporal factors in school closure policy for
  mitigating the spread of influenza,'' {\em Journal of public health policy},
  vol.~32, no.~2, pp.~180--197, 2011.

\bibitem{kim2012temporal}
H.~Kim and R.~Anderson, ``Temporal node centrality in complex networks,'' {\em
  Physical Review E}, vol.~85, no.~2, p.~026107, 2012.

\bibitem{Soh2014a}
H.~Soh and Y.~Demiris, ``Spatio-temporal learning with the online finite and
  infinite echo-state gaussian processes,'' {\em IEEE Transactions on Neural
  Networks and Learning Systems}, vol.~26, June 2014.

\end{thebibliography}

\end{document}